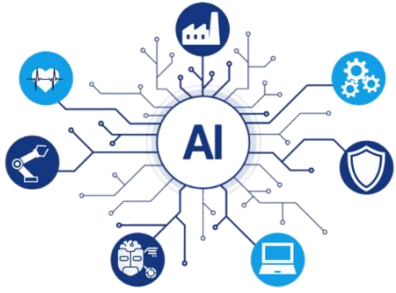

# Analysis of Filter Size Effect In Deep Learning

Yunus Camgözlü[1], Yakup Kutlu[2]

[1]The Graduate School of Engineering and Science, Iskenderun Technical University, Turkey
[2]Department of Computer Engineering, Iskenderun Technical University, Turkey
Emails: yunuscamgozlu@gmail.com, yakup.kutlu@iste.edu.tr

**Abstract**

With the use of deep learning in many areas, how to improve this technology or how to develop the structure used more effectively and in a shorter time is an issue that is of interest to many people working in this field. Many studies are carried out on this subject, it is aimed to reduce the duration of the operation and the processing power required, except to obtain the best result with the changes made in the variables, functions and data in the models used.

In this study, in the leaf classification made using Mendeley data set consisting of leaf images with a fixed background, all other variables such as layer number, iteration, number of layers in the model and pooling process were kept constant, except for the filter dimensions of the convolution layers in the determined model. Convolution layers in 3 different filter sizes and in addition to this, many results obtained in 2 different structures, increasing and decreasing, and 3 different image sizes were examined. In the literature, it is seen that different uses of pooling layers, changes due to increase or decrease in the number of layers, the difference in the size of the data used, and the results of many functions used with different parameters are evaluated.

In the leaf classification of the determined data set with CNN, the change in the filter size of the convolution layer together with the change in different filter combinations and in different sized images was focused. Using the data set and data reproduction methods, it was aimed to make the differences in filter sizes and image sizes more distinct. Using the fixed number of iterations, model and data set, the effect of different filter sizes has been observed.

## 1. INTRODUCTION

In today's technology, many branches of artificial intelligence serve people in different ways. Deep learning, one of these branches, is also used for different purposes in many fields of study. Deep learning is used in many areas such as face recognition, classification based on various data, healthcare and autonomous vehicles. Deep learning performs operations such as classification and object recognition by learning different features of input data with different algorithms such as CNN, RNN, PNN, DBN, due to its multi-layer structure. When deep learning is used for different jobs in many fields, a detailed research and test processes based on this research are reached in order to reveal the best results in the fastest way. The purpose of this approach is to achieve higher performance or higher efficiency in less time while optimizing the systems created, and to support later studies. Convolutional neural network, which is a deep learning algorithm, enables to perform the desired operation by learning the properties from large amount of data with its multi-layer structure.

When the literature is examined, it is striking that the studies are scarce both in general and in the field of leaf classification. In general, many studies such as filter number, number of layers, comparison of different models and different functions stand out in the studies. Current studies focus more on the filter size in the pooling layer.

Considering the studies examining the situations that occur according to the change of filter size, the optimum values in the parameters were obtained according to the results obtained by changing the image sizes used as input, changing the filter size, filter number and layer number (Sinha, 2017). The filter size was evaluated according to the results obtained as a result of how long it will be trained during the iteration and whether many different functions are used or not (Pasi, 2016). Results were obtained by comparing different models with different functions and values (Mishkin, 2016). Specifically, the filter size is set as a continuous variable, thus proposing a new method in which filter sizes and weights are learned simultaneously from training data (Han, 2017). The changes in the convolution layer filter values and the situations that occur as a result of the input values being composed of different numbers of dimensions have been examined (Du, 2018).

This study focuses on the change in the filter dimensions of the convolution layers in the model determined while the leaf is classified with CNN of the data set consisting of the leaf images with a fixed background, the difference of the filter sizes changing by increasing or decreasing, and the change to be obtained in different image sizes. Leaf images in the data set were reproduced with data replication processes, converted to gray images and prepared to be used in 3 different sizes. Then the model and a fixed number of iterations are determined. Filter sizes were determined as 3, 5 and 9, in addition, a structure where the filter size increased from 3 to 5, then 9, and another structure where the filter size was reduced from 9 to 5 and then to 3. The tests were repeated 3 times for 66 x 50, 90 x 75 and 132 x 100 sized images, and the results are presented in this study as an average.

The study aimed to examine the effects of different filter sizes with many changeable structures such as fixed number of layers, fixed iteration and fixed variables in different sized images.

2. **MATERIALS AND METHODS**

2.1. **Database**

After examining many data sets used in the literature and conducting various experiments, the results were obtained from 6 different data sets, independent of scaling, only rotation was applied, and the results were not suitable for the 9x9 dimensional filter when testing different filter sizes.

In this case, based on 12 types of leaf species image consisting of 4404 units on a fixed background data set was decided to use Mendeley data set (Chouhan 2019). In this data set, the data replication process has been applied in order to have a small number of images and to make the results more accurate. A total of 52624 images were obtained for 12 types, together with the original images and the images obtained by rotating the images of each type in 11 different angles.

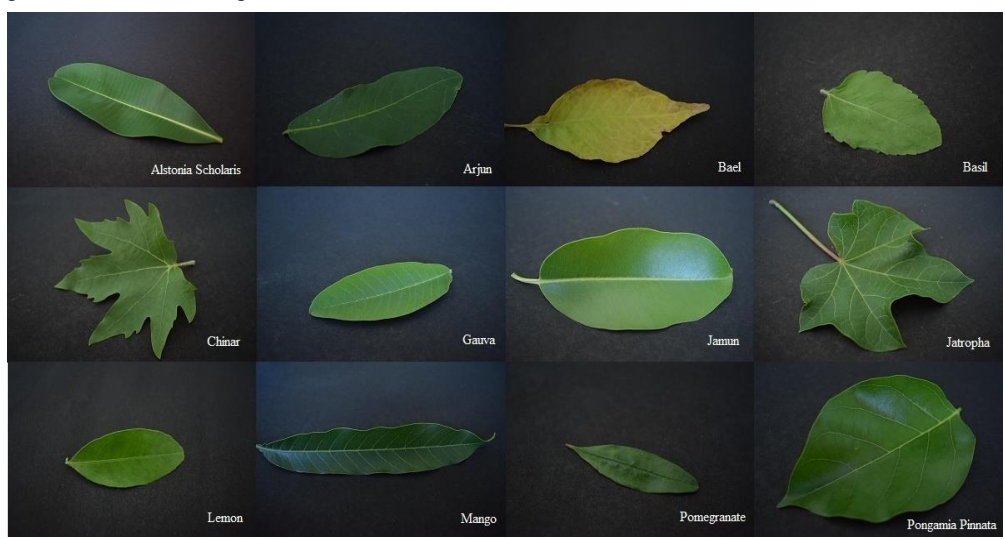

**Figure 1.** Sample image of each leaf species in the dataset

*Corresponding Author: Yakup Kutlu, E-mail: yakupkutlu@gmail.com*

This data set, which includes leaf images of different sizes and close-colored backgrounds of these images, also includes images of diseased leaves. The uneven images were removed from the data set, since the performance would decrease considerably during the classification before pretreatment.

In the data duplication process made to the data set used, the images were rotated over the angle values starting from 30 degrees to 330 degrees with 30 degrees increments. Together with the original image, there are 12 different views of each leaf image in total. These images were created in 3 different sizes, 66x50, 90x75 and 132x100.

**2.2. Convolutional Neural Networks**

The CNN structure is the combination of three ideas: 1) Local receptive fields, 2) shared weights, and 3) sub-sampling. Local receptive fields force CNN's to extract features since each particular layer comprises several feature maps (Risueno, 2002). In addition to many layers such as convolution, pooling, leveling and classification, these layers are used in functions with various functions. These layers and functions are used in different numbers and orders, and various models are created to obtain the best result according to the available data.

Convolution layer has critical importance for CNN's operations such as classification and definition. Feature maps are created with different size filters and different number of convolution layers applied. As a result of the operations performed, the data also shrinks, in the face of this event, the filling process is applied to the edges of the picture in order not to lose the information on the edges of the picture, and a better and faster result is obtained by using various functions. The pooling layer that stands out between convolution layers exists in two different types. These are maximum pooling and average pooling. In this layer, the filter performs the process by taking the maximum value or the average value in the area where the operation is carried out by moving over the data.

The leveling layer after a series of convolution and pooling layers transmits the incoming multidimensional matrix to the classification layer in the form of a one-dimensional matrix. Thanks to the high level properties revealed by the previous layers in the one-dimensional matrix coming from the plane layer, the training is carried out in the classification layer with the neural network in FC. Neurons in the neural network examine the paired value and label pair in the one-dimensional matrix according to each label. Then, labels with high accuracy are determined.

**3. The Model**

Different number of pooling were used to show pooling effect in leaf classification with CNN. In addition two pooling method, which are max pooling and average pooling, were used in this study (Camgözlü, 2019). In line with the results obtained as a result of the study, the convolution number was chosen as 6 and the pooling number as 3. In the pooling method, average pooling was used, and softmax was chosen for classification.

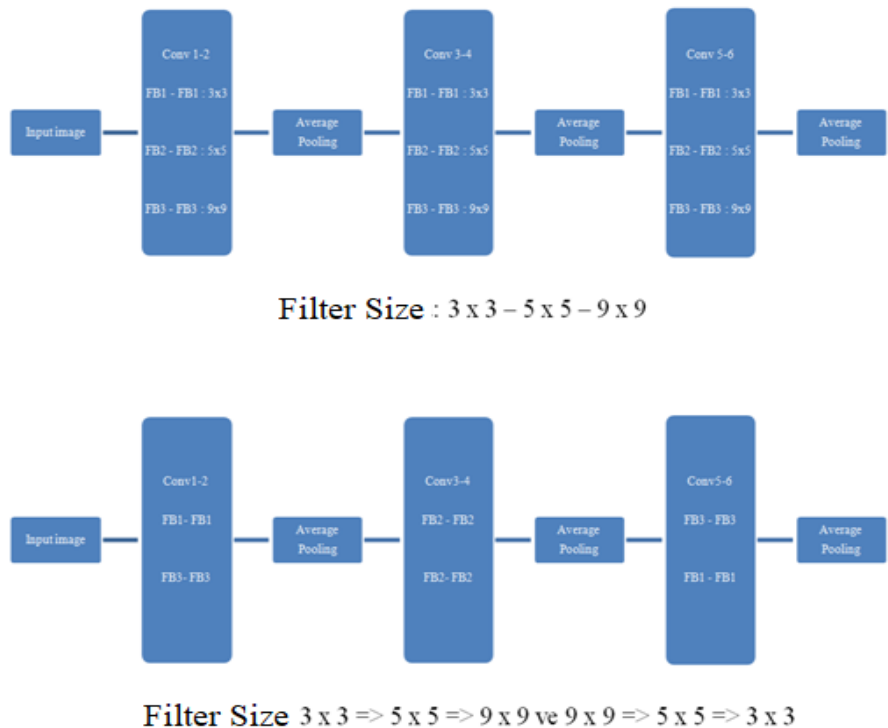

**Figure2.** Display of different filter sizes used in the mode (Camgözlü, 2019

*Corresponding Author: Yakup Kutlu, E-mail: yakupkutlu@gmail.com*

In order to examine the filter sizes in the convolution layer, which is the aim of this study, the filter sizes of the convolution layers in the model were arranged in 3 different values and 3 different ways, and the differences caused by these changes were examined. Each of the 6 convolution layers has 3, 5 and 9 filter sizes. In order to determine the effect caused by the differences in filter sizes, 3 different filter sizes were used by creating separate models in each convolution.

In addition, in cases where the filter size increases and decreases, the filter sizes are increased or decreased after every 2 convolution layers.

## 3. RESULTS

The models created while performing operations such as classification and definition with CNN are of great importance. These models are likely to increase, decrease or change the duration of the operation according to many concepts such as data type, number, structure of the shape to be processed. In such cases, structures in the literature are examined and high-performance structures applied in similar problems are determined, and many trial trainings are carried out to create a different model from these approaches and a faster and better performance is tried to be obtained.

In this study, according to a model created in leaf classification with CNN of Mendeley (Chouhan, 2019) data set, the filter size in the convolution layer gradually increases or decreases in the filter size in the same model and the comparison of the fixed value filter size in the model created 3 different images. Made for size. With the pioneering tests, different filter sizes to be tested for this study, the number of iterations and various parameters were determined. In addition, the models created to determine the effect of the increase and decrease in filter size were run 3 times.

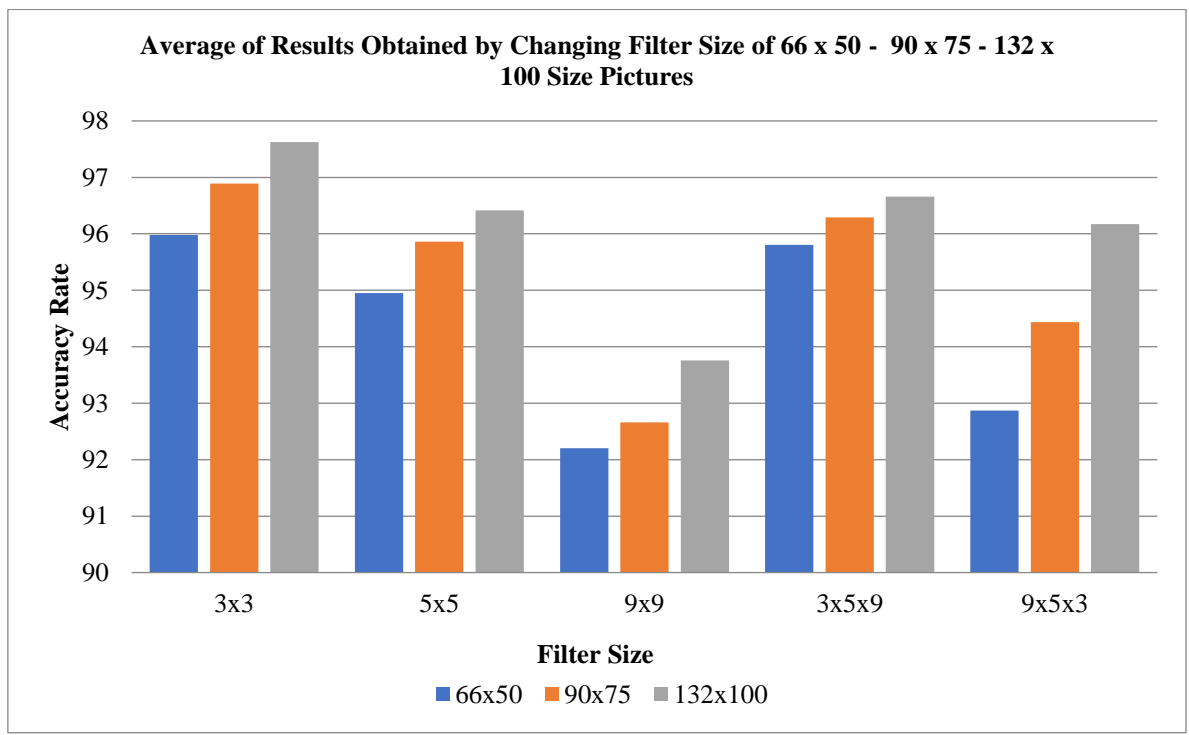

**Figure3.** The graph of average of the accuracy rates obtained as a result of the experiments

According to these results, it is seen that the increase in filter size in 3 different image sizes negatively affects the performance. In addition, it is seen that the increase of different filter sizes in different convolution layers in the same model decreases the performance very little, while the decrease of these filter sizes decreases the performance more.

Corresponding Author: Yakup Kutlu, E-mail: yakupkutlu@gmail.com

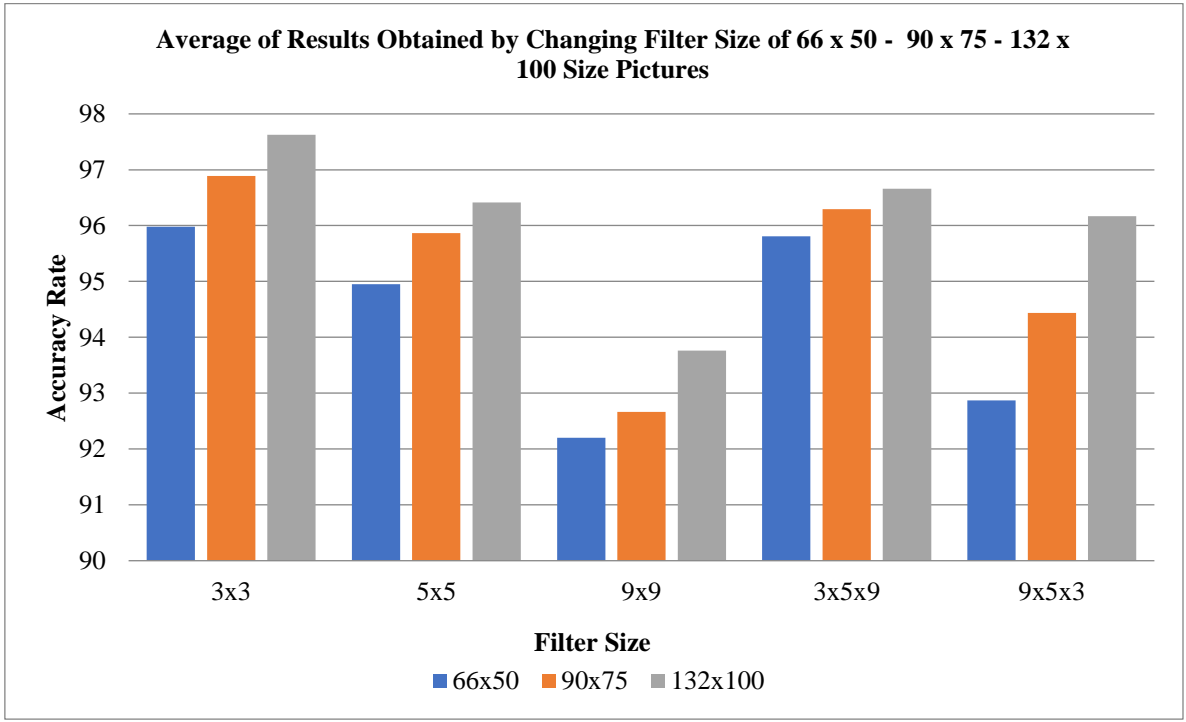

**Figure 3.** The graph of the average results of the accuracy rates obtained as a result of the experiments.

According to these results, it is seen that the increase in filter size in 3 different image sizes negatively affects the performance. In addition, it is seen that the increase of different filter sizes in different convolution layers in the same model decreases the performance very little, while the decrease of these filter sizes decreases the performance more.

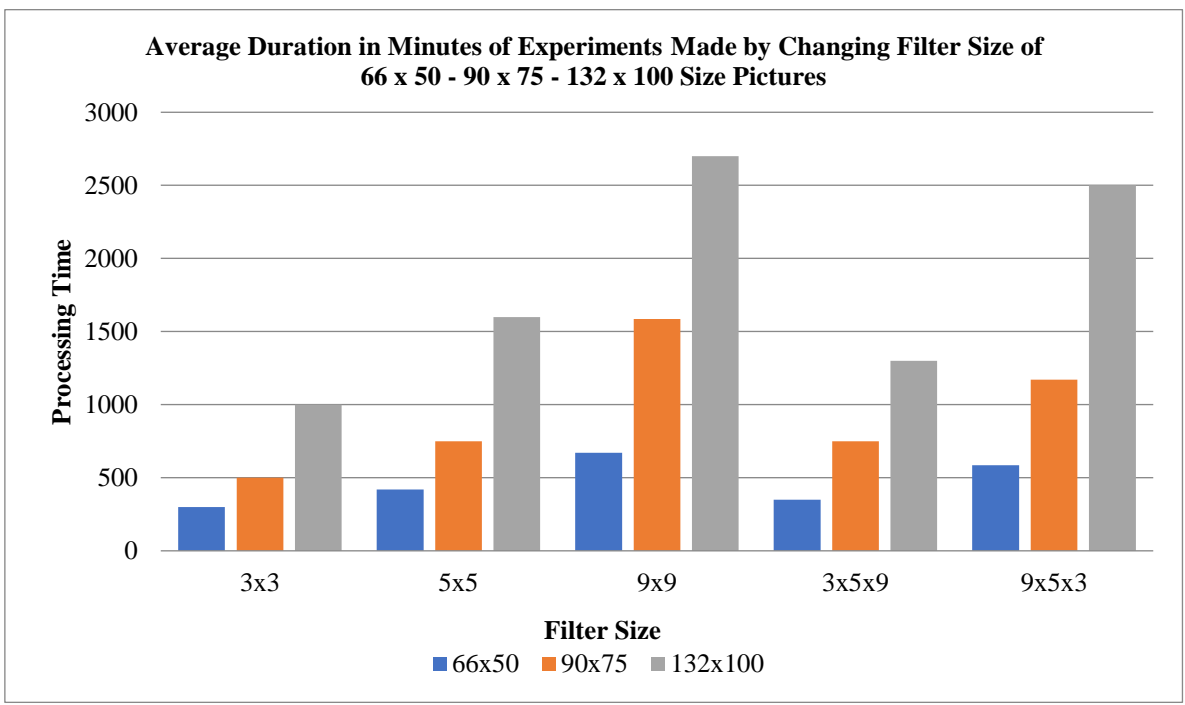

**Figure 3.** The graphic of the approximate duration in minutes of 66x50 - 90x75 - 132x100 sized pictures as a result of the experiments made with the filter size change

*Corresponding Author: Yakup Kutlu, E-mail: yakupkutlu@gmail.com*

In line with this graph, it is an expected result that the increase in image size and the increase in processing times is expected to be related, while looking at the graph showing the achievements, the change in performance rate for low filter size is about 1%. In addition, the effect of the changes in filter size on the process time reveals that it is more appropriate to choose the filter size. In addition, increasing the number of iterations instead of the image size by looking at the change in the performance rate can increase the performance more in lower performance situations.

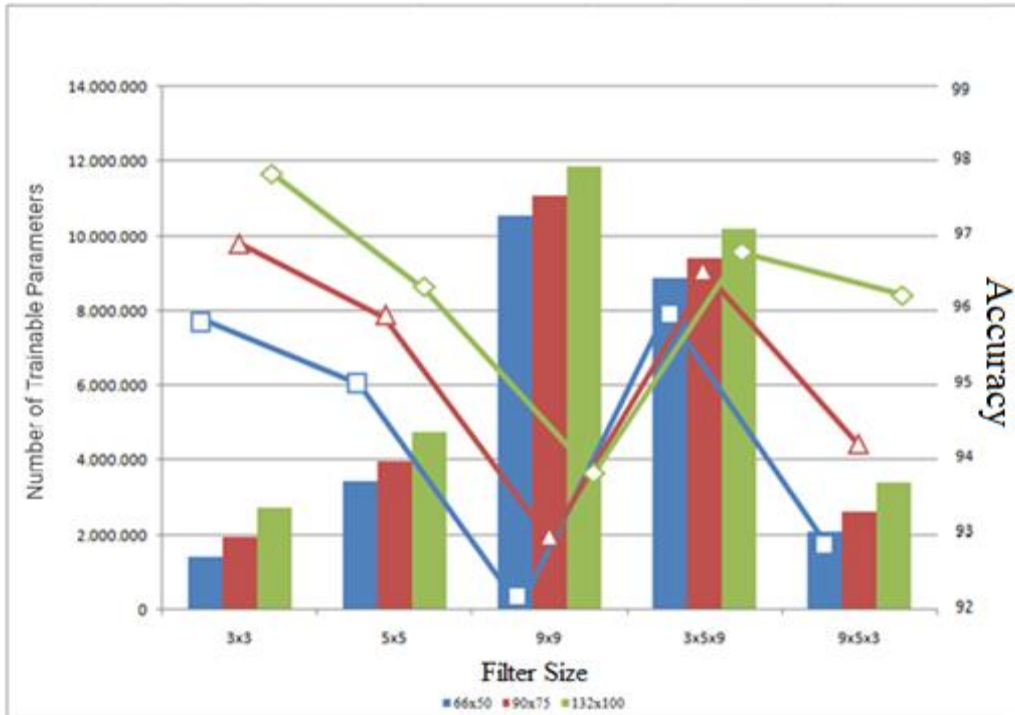

**Figure 4.** Trainable parameter number and performance ratio comparison chart for 66x50 - 90x75 - 132x100 sized images

When we look at the number of trainable parameters that vary according to the image size with the convolution filter size, and the time required for the training, it is seen that the increase in the filter size greatly increases the number of trainable parameters. As a result, it has been revealed that structures with more trainable parameters need to train with much more iterations in order to achieve higher performance. In structures where the filter size increases from low to high and decreases from high to low, the low initial filter size of the model contributed to the extraction of more trainable parameters by preventing the loss of more obtainable data. In addition, if we look at the success rates and the time elapsed while obtaining these success rates, it is seen that the trainable parameters obtained have increased as a result of the gradual increase of the filter size, and it is also seen that the filter size requires less processing time compared to the structure where the filter size decreases from high to low.

*Corresponding Author: Yakup Kutlu, E-mail: yakupkutlu@gmail.com*

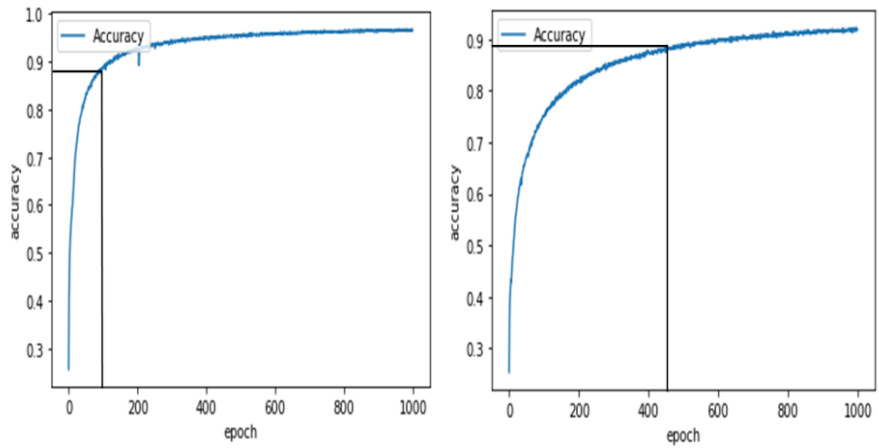

**Figure 5.** Performance ratio comparison of low filter sizes in 100 iterations for 3x3 and 9x9 filter sizes of 90x75 size images

When the number of trainable parameters are examined together with the performance charts selected from the performance charts obtained in the experiments for 3x3 and 9x9 filter sizes of 90x75 sized images, it is seen that achieving a lower success in 100 iterations is a result of the 6-fold difference in the number of trainable parameters. This difference is seen in the performance graph above as a difference of approximately 6 times in the number of iterations required to achieve the success achieved with low filter sizes at 100 iterations.

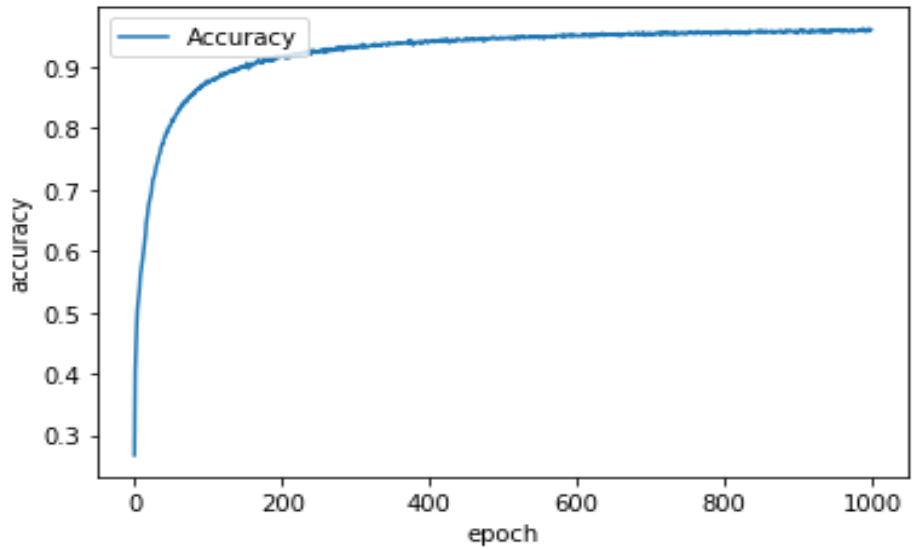

**Figure 6.** Graph showing the success achieved after 100 iterations for 3x3 filter size in 66x50 images

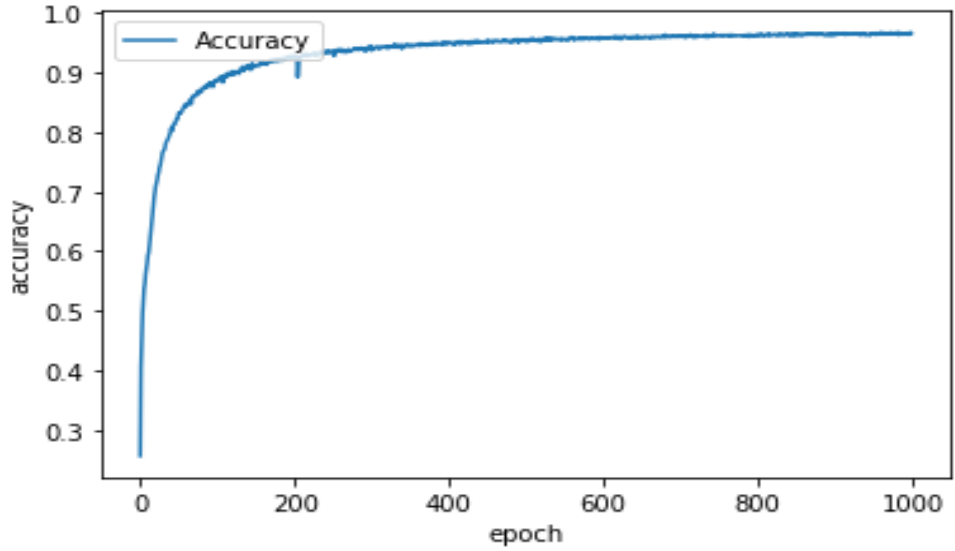

**Figure 7.** Graph showing the success achieved after 100 iterations for 3x3 filter size in 90x75 images

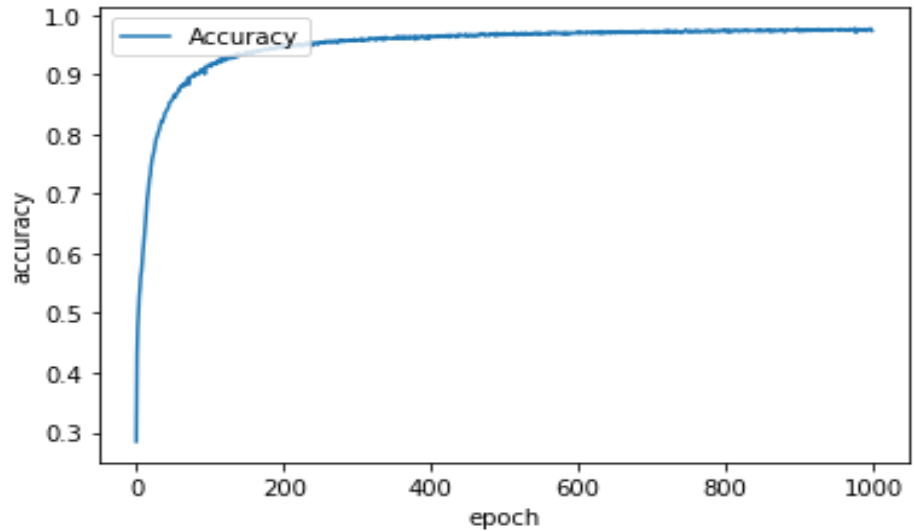

**Figure 8.** Graph showing the success achieved after 100 iterations for 3x3 filter size in 132x100 images

When we look at the performance curve achieved by 3x3 convolution filter size in 3 different image sizes at 100 iterations, it seems to be close to each other, although it provides the best performance of 132x100 sized images when we consider the processing time, it requires 2 times more time in terms of processing time.

Considering that 66x50 sized images may not always give good results in different data sets and studies to be carried out in different fields, and it has been observed that the performance increases as the image size increases, it has been determined that 90x75 sized images are better for general us.

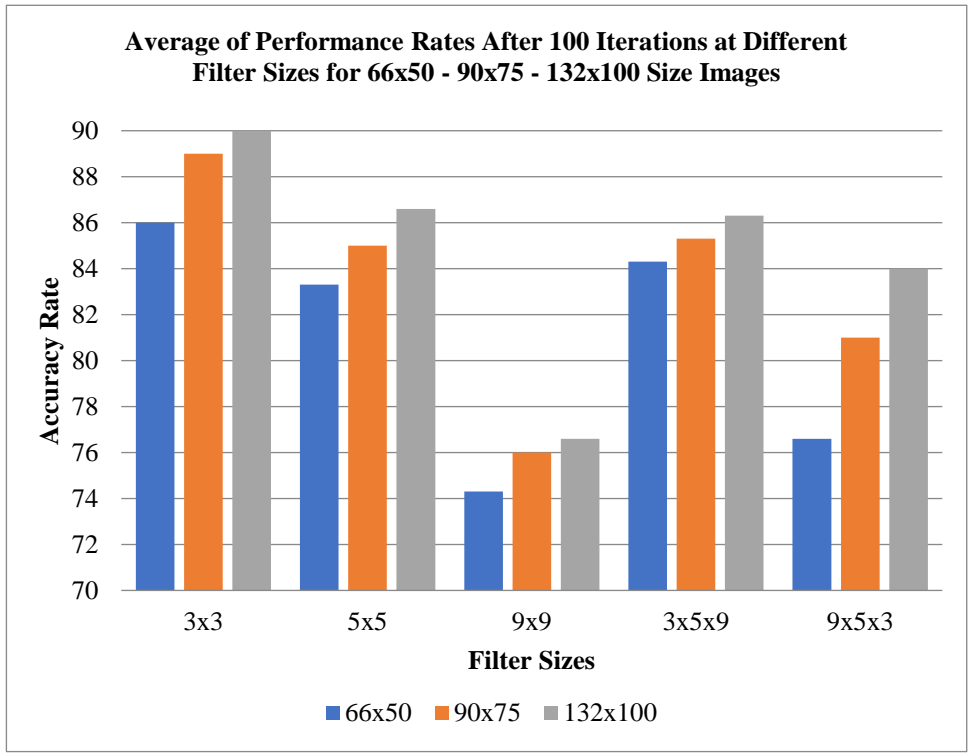

**Figure 9.** Graph showing average of performance rates obtained after 100 iterations in different filter sizes for 66x50 - 90x75 - 132x100 sized images

When we look at the success achieved by 100 iterations for 3 different images, it is seen that 3x3 filter size achieved close results. When we look at the required times and performance rates for these processes in detail, it is seen that the lower filter size exhibits better performance and when we consider the less processing time brought by the lower filter size, the l

### 4. CONCLUSION AND DISCUSSION

According to the results obtained as a result of the study, the reduction of the filter size provides both an increase in performance, a shorter time to reach success and a shorter duration of the process as a result of the low filter size. While it is expected that the performance will increase as the image size increases, the processing times obtained are of great importance in evaluating these performance values. It is an indisputable fact that processing time is critical, especially in situations that are made with low hardware or require a lot of repetition.

As a result of the experiments conducted in this study, the increase in image size brings little increase in performance for the low filter size, and the fact that it requires 2 times more time than the situation in processing times caused this option to be eliminated. In the lower image size, the more data lost during the downsizing stage caused the performance not to increase or to require more iteration in order to increase it.

lower filter size is more appropriate. In addition to this, the advantage of starting with a low filter size in the structure in which the filter size is increased, better results have been obtained as a result of the more data that can be used in the subsequent layers after the application of less reduction of the filter applied than the 9x5x3 filter sizestructure

Choosing a low filter size of 90x75, which we can call the middle way, provides a great gain in processing time compared to higher image sizes. In addition, it is predicted that the performance loss obtained can be compensated by an iteration increase of 10% - 20%.

In order to expand the study, it is necessary to try more filter sizes and different combinations in different data sets. In addition, in order to better understand the change in filter size, a data set with too many species and too many leaf images can be considered in a different study that can be done by gradually reducing the number of species and images.

*Corresponding Author: Yakup Kutlu, E-mail: yakupkutlu@gmail.com*